\documentclass[journal]{IEEEtran}
\ifCLASSINFOpdf
   \usepackage[pdftex]{graphicx}
\else
\fi
\usepackage{amsmath}
\usepackage{amssymb}

\usepackage{array}

\usepackage{stfloats}
\begin{document}
%
\title{Bi-directional Graph Structure Information Model\\ for Multi-Person Pose Estimation}
%
%
%

\author{Jing~Wang*, Ze~Peng*, Pei Lv, Junyi~Sun, Bing~Zhou
        and~Mingliang~Xu
\thanks{Jing Wang, Pei Lv, Ze Peng, Junyi Sun, Bing Zhou and Mingliang Xu
are with Center for Interdisciplinary Information Science Research, ZhengZhou
University, 450000. {iexumingliang, ielvpei, iebzhou}@zzu.edu.cn;
wangjingzzu@gs.zzu.edu.cn; {jysun, zpeng}@ha.edu.cn. $*$ represents Jing Wang and Ze Peng contribute equally to this study. Pei Lv is the corresponding author.}
}

\maketitle

\begin{abstract}
The performance of Multi-person pose estimation has been greatly improved in recent years, especially with the development of convolutional neural network. However, there still exists a lot of challenging cases, such as crowded scene and ambiguous overlapping, which cannot be well addressed. In this paper, we propose a novel bi-directional graph structure information model (BGSIM) for multi-person pose estimation to relieve above problems. More specifically, the BGSIM is contained in a multi-stage network architecture with two branches. The first branch predicts the confidence maps of joints and uses a geometrical transform kernel to propagate information between neighboring joints at the confidence level. The second branch leverages the BGSIM to encode rich contextual information and to infer the occlusion relationship among different joints. In the BSSIM, we dynamically determine the joint point with highest response of the confidence maps as the base point of passing message. The base point propagates information to other joints (especially for the occluded joints) via unrolled tree structures, this ensures reliable information which is transmitted among different body parts and facilitates the prediction of joints for multi-person. Based on our proposed model, we achieve an average precision of 62.9 on the COCO Keypoint Challenge dataset and 77.6 on the MPII (multi-person) dataset. Moreover Compared with other state-of-art methods, our method can achieve highly promising results on our selected multi-person dataset without extra training.
\end{abstract}


%
\IEEEpeerreviewmaketitle

\section{Introduction}
%
%
%
%
\IEEEPARstart{H}{uman} pose estimation from still image is a challenging problem in computer vision. It is key to many visual tasks, such as action recognition~\cite{Song2011Localized,Yu2015Propagative,Zhou2008Activity}, person re-identification~\cite{Zhu2017Deep}, and human-computer interaction~\cite{Jain2011Real}. Human pose estimation contains single-person pose estimation and multiple-person pose estimation and recognizing the poses of multiple persons is a lot more challenging than recognizing the pose of a single person in an image~\cite{Ladicky2013Human,Newell2016Stacked,Sapp2010Cascaded,Sun2012Conditional,Wei2016Convolutional}. So many researchers are still focus on this problem.

Recently, significant advancements have been achieved in this research field by deep convolutional neural networks (DCNN)~\cite{Kang2016Object,Chen2016Once}. Some works~\cite{Papandreou2017Towards,Huang2017A,He2017Mask,Fang2016RMPE} use human detectors to obtain all the bounding boxes of individuals in an image and then perform single-person pose estimation in each of the detected bounding boxes. However, these top-down approaches do not work very well in complex or crowded environments, where individuals cannot be detected accurately. Meanwhile, other researchers use bottom-up approaches~\cite{Cao2016Realtime,Newell2016Associative,Insafutdinov2016DeeperCut,Pishchulin2015DeepCut} directly predict all joints and assemble them into different individuals. Although great progress has been made by this kind of method, there still exist a lot of challenging cases, such as crowded scene, self occlusion and other occlusion, which cannot be well predicted. The main reasons lie at two points: 1)the joints are wrongly connected due to too many people, 2)there exists some ambiguous overlapping when a joint of an individual is obscured by the same one of another individual.

To address above problems, we propose a novel bi-directional graph structure information model (BGSIM) for multi-person pose estimation. The BGSIM is contained in a network architecture which has three stages. In our method, we don't directly predict all body joints, we divide the full body pose estimation into three subsets \{head, shoulders\}, \{elbows, hips\} and \{wrists, knees, ankles\}. So we first predicts confidence maps of head and shoulders in first stage, because heads and shoulders are easier to predict among all body parts due to their relatively fixed position. Then the elbows and hips with relatively flexibility are predicted in the second stage. The third stage will consider the full body joints. Each stage has two branches in our network architecture. The first branch learns confidence maps of each body joint based on basic feature which is captured by the VGG~\cite{Simonyan2014Very}. More importantly, this part also implements a geometrical transform kernel~\cite{Chu2016Structured} with a convolution layer to shift the response between neighboring joints of a person at the confidence map level. Compared with those work on feature level~\cite{Chu2016Structured,Yang2017Learning}, we can capture more explicit location information of the joints.

 \begin{figure}[t!]
\centering
\includegraphics[width=3.2in]{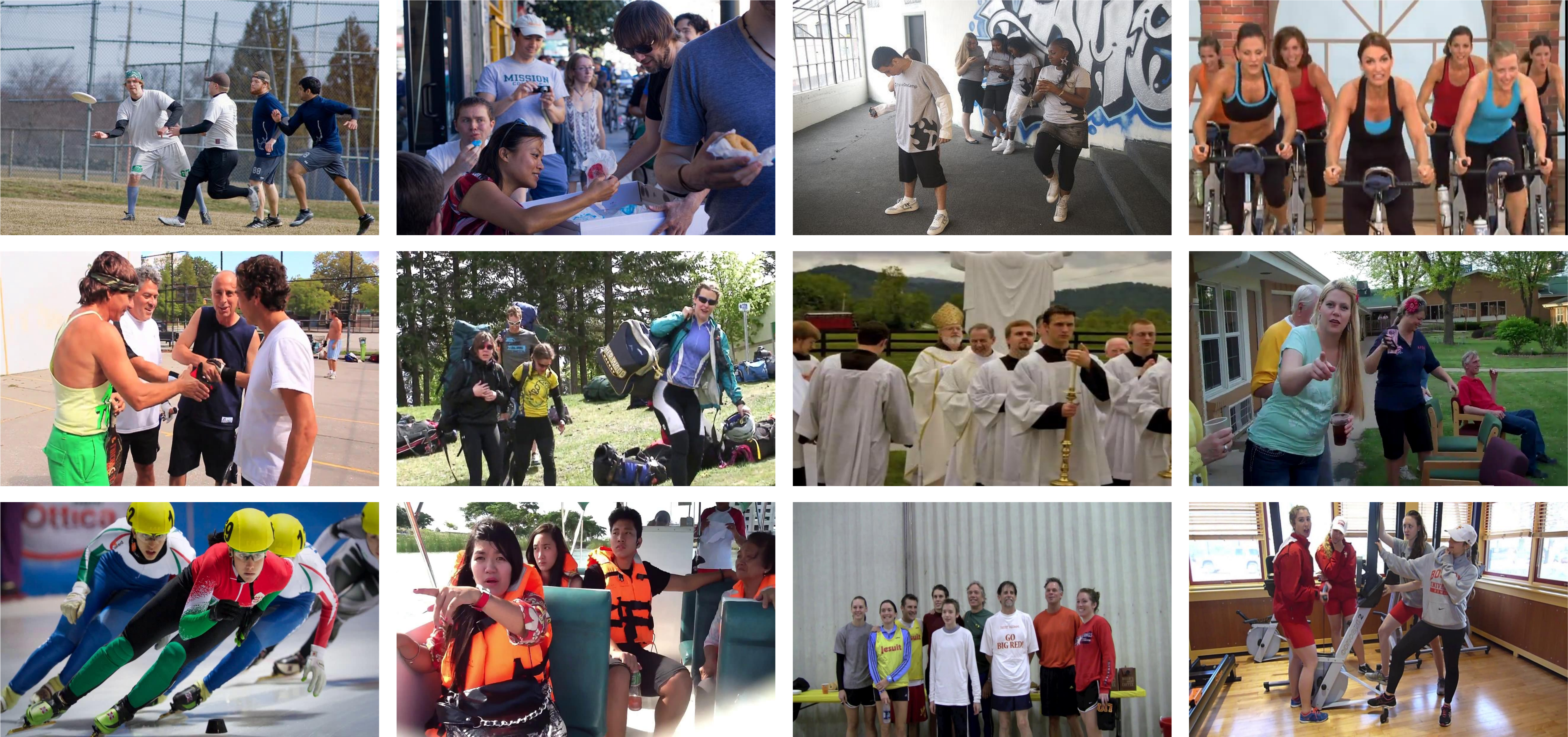}
\caption{Some image samples from our selected multi-person dataset. These images containing crowded people and high occlusion are chosen from the AI challenger, COCO, MPII, and LSP public datasets}
\label{fig:1}
\end{figure}

Based on the confidence maps estimated in the first branch, the second branch BGSIM explicitly address those occluded joints. Different from traditional tree structure model, BGSIM is a graph structure model for multi-person pose estimation. It can model the interactions between unconnected body parts, and the interactions between human body and objects. These interactions are important cues for occlusion reasoning to capture rich contextual information. As the proposed BGSIM is a graph containing loops, we perform inference on tree structures unrolled by the BGSIM. Because one joint point with highest response of confidence maps contains most reliable information, we dynamically determine it as the base point in the BGSIM. And the base point propagates information to other joints (especially for the occluded joints) via the tree structures. This ensures we can encode richer contextual information containing some invisible and occluded joints to achieve good performance for multi-person pose estimation.

Our proposed method achieves state-of-the-art results in the COCO and MPII benchmark datasets. We also construct a new dataset that contains 5899 images of multiple people with different degrees of occlusions. Some image samples are shown in Figure~\ref{fig:1}. On this dataset, our method can achieve a greater improvement in accuracy compared with the existing methods.

The contributions of our work are summarized as follows:
\begin{itemize}
\item We propose a BGSIM to encode rich contextual information and to infer occlusion relationship among different body parts.
\item We determine the base point dynamically in BGSIM and use it as the root node for unrolled tree structures to propagate information to other joints.
\item We involve the BGSIM into a network architecture with three stages to capture more explicit location information of body joints at the confidence level for multi-person pose estimation.
\end{itemize}



%
%

\section{Related Work}

Automatic human pose estimation is crucial to intelligent surveillance systems, especially for human tracking and action recognition. Image-based human pose estimation in real-world environments remains a challenging task, and its current performance is still far from satisfactory due to occlusion and multi-person problems in complex scenes.

These problems are usually solved by using pictorial structures~\cite{Fischler1973The,Andriluka2009Pictorial} or graphical models~\cite{Chen2014Articulated}. The Pictorial Structure Model~\cite{Andriluka2009Pictorial} defines some pairwise connections of independent parts to represent the correlations among different body parts. However, the tree-structured model is ``oversimplifie''. Accordingly, many researchers proposed highly complex tree-structured models and the graph structure model~\cite{Fu2017ORGM}. Tian et al.~\cite{Tian2012Exploring} proposed a hierarchical spatial model that captured an exponential number of poses with a compact mixture representation on each part. L. Pishchulin et al.~\cite{Pishchulin2015DeepCut} proposed a typical approach to articulate pose estimation by combining the spatial modeling of the human body with the appearance modeling of body parts. Meanwhile, other approaches~\cite{Tian2012Exploring,Dantone2013Human,Gkioxari2013Articulated,Pedro2005Pictorial,Sapp2013MODEC,Wang2013Beyond,Wang2011Learning,Sapp2010Adaptive,Pishchulin2013Poselet} used the tree-structured or graphical model to solve the problems of human keypoints estimation and predicted the keypoint locations based on handcrafted features (e.g., SIFT and HOG). With the rapid development of deep convolutional neural networks (CNN), recent studies~\cite{He2017Mask,Chu2016Structured,Toshev2013DeepPose,Jain2014MoDeep,Kang2016Object,Ouyang2016DeepID,Bulat2016Human,Gkioxari2016Chained,Wei2016Convolutional,Yang2017Learning} have made great process in this research area.

\textbf{Single-Person Pose Estimation} \ Early studies on human pose estimation mainly focused on single-person pose estimation due to its simplicity. Toshev et al. introduced CNN to solve the pose estimation problem in DeepPose~\cite{Toshev2013DeepPose}, which cascaded CNN pose regressors to deal with the body joints. Wei et al.~\cite{Wei2016Convolutional} and Newell et al.~\cite{Newell2016Stacked} found that deep CNNs performed effectively in human pose estimation as well as proposed a multi-stage architecture to solve the pose estimation problem. Newell et al. compiled several hourglass modules and proposed a U-shape architecture for human pose prediction. X. Chu et al.~\cite{Chu2016Structured} designed a structured feature-learning architecture to infer the correlations among different body parts at the feature level for human pose estimation. Although CNN can effectively extract feature information, the occlusion relationship among different body parts have not been fully explored.

\textbf{Multi-Person Pose Estimation} \ The above approaches show a promising performance in single-person pose estimation, but real-world images usually contain multiple persons. Multi-person pose estimation is a challenging task due to occlusion, the unpredictable interactions among different persons, and the diversity of human gestures. Some researchers applied bottom-up approaches to deal with this problem. For instance, DeepCut~\cite{Pishchulin2015DeepCut} directly predicted all body part candidates, categorized them into different classes, and partitioned them into the corresponding people. DeeperCut~\cite{Insafutdinov2016DeeperCut} employed image-conditioned pairwise terms to achieve better pose estimation performance based on ResNet~\cite{He2016Deep}. Cao et al.~\cite{Cao2016Realtime} mapped the relationship among different keypoints into part affinity fields (PAFs) and assembled the detected keypoints into different poses of people. In contrast to bottom-up approaches, top-down approaches initially located each person and then independently analyzed their joints. For example, Mask-RCNN~\cite{He2017Mask} initially predicted the human bounding boxes, and then croped out the feature map of these bounding boxes to predict the human keypoints. Papandreou et al.~\cite{Papandreou2017Towards} predicted both the dense heatmaps and offsets of each keypoint type from each bounding box by using a fully convolutional resnet, and then fused these heatmaps with offsets to obtain the final predicted location of keypoints. RMPE~\cite{Fang2016RMPE} predicted human pose by using the stacked hourglass model and then adjusted the imperfect proposal by combining STN, SPPE, and SDTN to overcome the location errors caused by the detection.

Nevertheless, these methods have not good enough performance in cases of multiple targets or serious occlusion. To address such limitations, we propose the BGSIM in our network architecture that can fully explore the connections among different body parts.
\begin{figure*}[t!]
\centering
\includegraphics[width=6.6in]{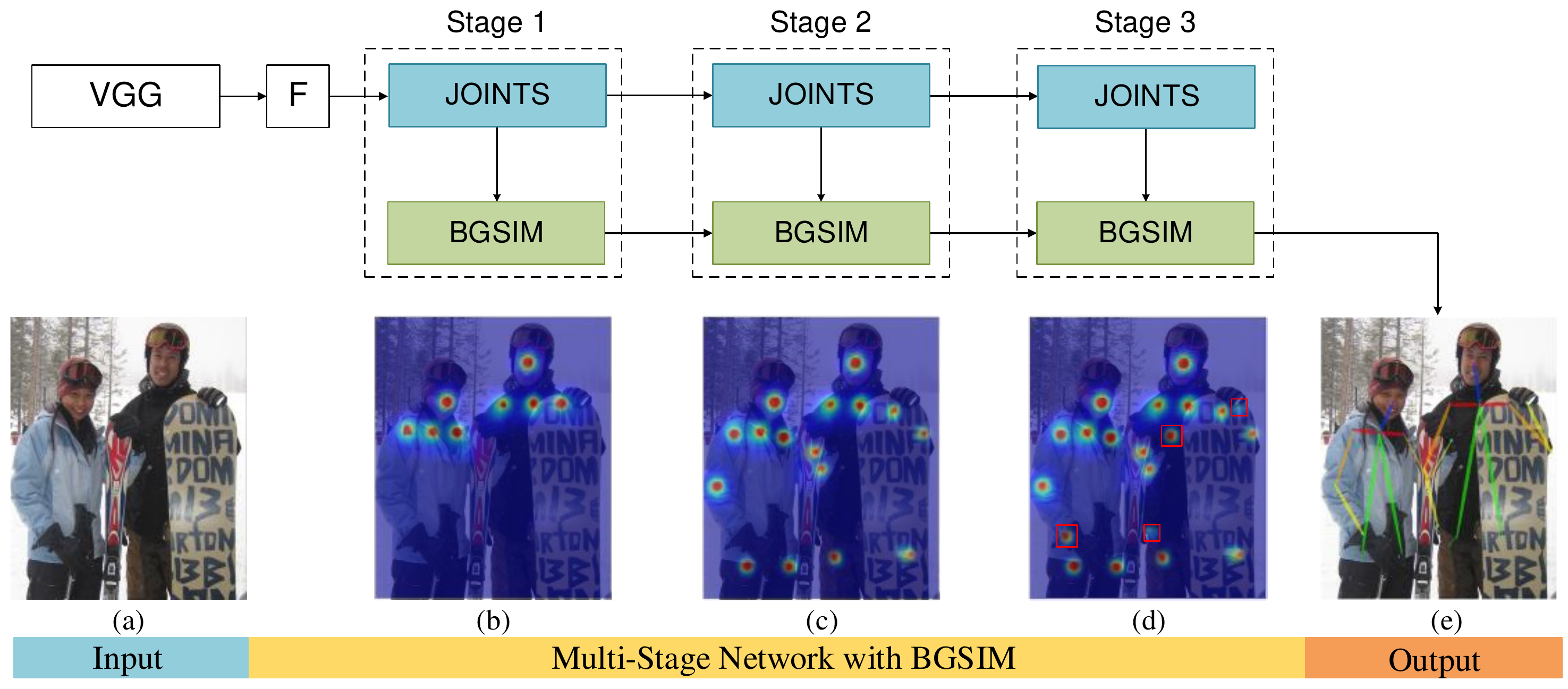}
\caption{Overall pipeline of our method. It takes the entire image as the input (Figure~\ref{fig:2} (a)); The multistage CNN network contains three stages and each stage has two branches JOINTS and BGSIM; The correct poses are eventually obtained for all persons (Figure~\ref{fig:2} (e)), the red square mark in Figure~\ref{fig:2} (d) represents the new confidence maps of occluded joints obtained in the third stage}
\label{fig:2}
\end{figure*}

\section{Our Method for Multi-person Pose Estimation}

Figure~\ref{fig:2} illustrates the overall pipeline of our method. A color image containing two persons is taken as an input and the accurate locations of keypoints for each person in the image is eventually obtained.  First, The input image is analyzed by a convolutional network initialized with the first 10 layers of VGG-19~\cite{Simonyan2014Very}. This convolutional network generates the basic feature information $F{\rm{ = }}g(I)$, where $I$ denotes the input image and $g$ is a nonlinear function.  Then, we use a multi-stage network architecture to handle the basic feature information $F$. The network has three stages, and each stage contains two branches: JOINTS and BGSIM. JOINTS predicts the confidence maps of joints. More specifically, we use a convolutional network to predict basic confidence maps of joints $c(x) = p(x|\sigma )$ of body part locations as defined in ~\cite{Zhang2013Fast}, where $x \in {R^2}$ denotes the joint location and $\sigma$ denotes the types of joints. BGSIM can explicitly address some occluded joints based on the confidence maps $c(x)$. If we are unable to capture information about heads or shoulders in the first stage, the second and third stage are still effective. The BGSIM can infer the information about heads and shoulders by using existing information of elbows and hips which is obtained in the second stage. More details about JOINTS and BGSIM will be described in Section~\ref{cm-sec} and Section~\ref{bsim-sec}.

\begin{figure}[t!]
\centering
\includegraphics[width=3.2in]{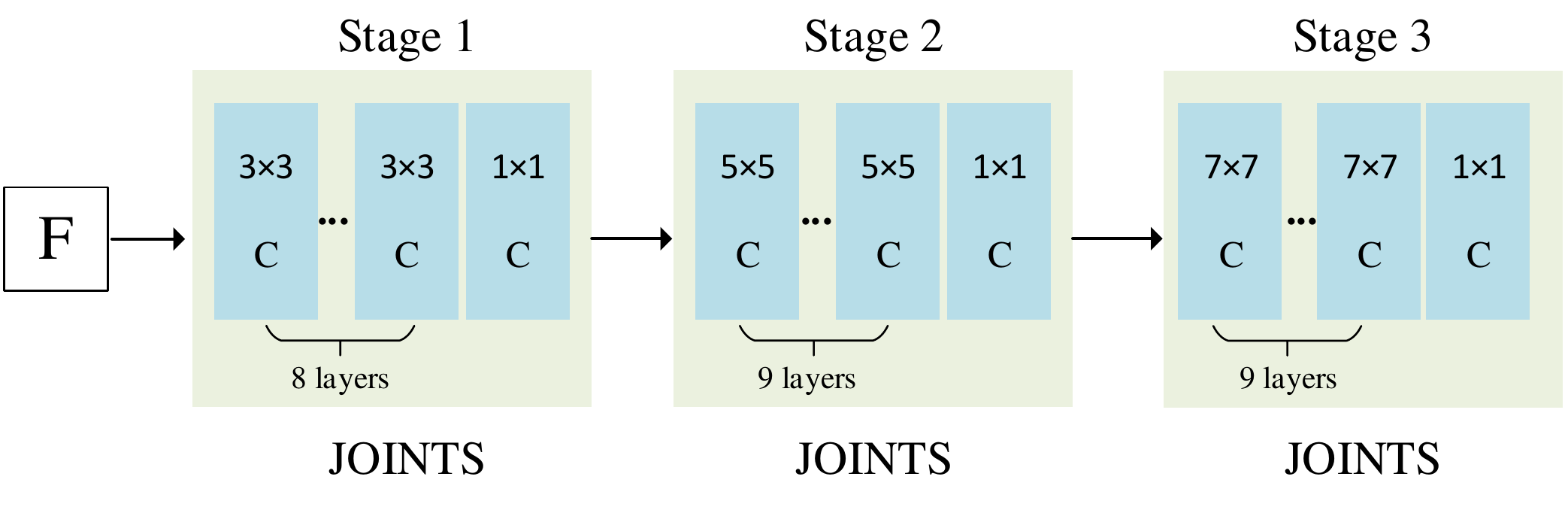}
\caption{The specific convolution parameters of JOINTS at each stage}
\label{fig:joint}
\end{figure}

\subsection{JOINTS}
\label{cm-sec}
As mentioned above, we need a reliable body part detection model to detect confidence maps of all the joints of a person before analyzing his or her pose. We use a deep, fully-convolutional human body part detection model that draws on powerful ideas regarding semantic segmentation, object classification, and human pose estimation.

Our convolutional network model is initialized with the first 10 layers of VGG-19~\cite{Simonyan2014Very} and is finetuned by changing the parameters of the remain layers (the names, number of output, and coefficient of learning rate). The branch JOINTS in the network architecture is to predict confidence maps of joints. It also has three stages, and the convolution parameters of JOINTS is different at each stage. Figure~\ref{fig:joint} shows specific convolution parameters of JOINTS at each stage. All 1*1 kernel size at each stage is to reduce dimension. From Figure~\ref{fig:joint} we can see that the deeper the number of stage, the larger the convolution kernel size, because the bigger kernel size can obtain bigger receive field which is helpful to predict the position of joints.

\begin{figure*}[t!]
\centering
\includegraphics[width=6.6in]{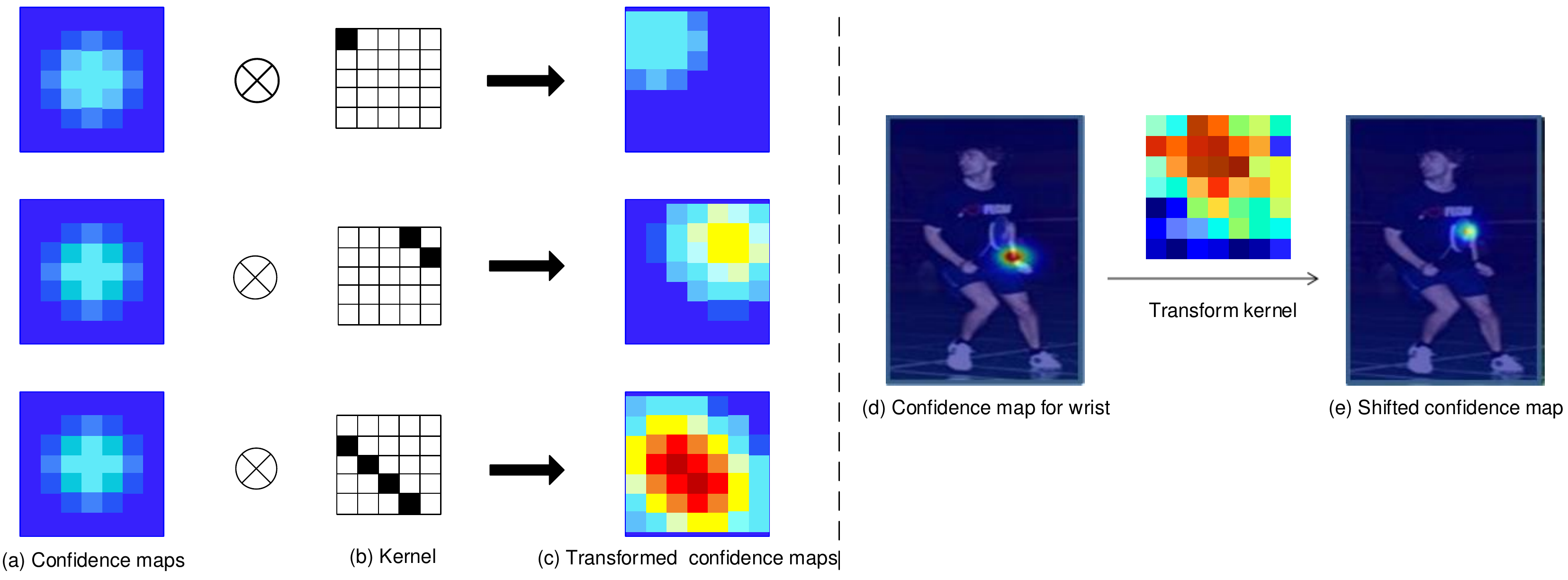}
\caption{(a)-(c) show that the confidence maps can be shifted through convolution with kernel. (a) is a confidence map assuming Gaussian distribution, (b) are different kernels for illustration, (c) are the transformed confidence maps after convolution. while (d)-(e) show an example of confidence maps that are shifted by geometrical transform kernel between adjacent joints}
\label{fig:4}
\end{figure*}

In addition, for JOINTS, we implement geometrical transform kernel by a convolutional layer. We can use geometrical transform kernel to shift confidence maps of neighboring joints. In order to illustrate the process of information passing, an example is shown in Figure~\ref{fig:4}. The confidence maps of elbow and wrist are estimated by the convolutional network and the confidence map of wrist has high response(as shown in Figure~\ref{fig:4}(d)). However, the elbow is occluded by other object, we cannot obtain its confidence map directly. It is not suitable to shift confidence map of wrist to elbow, since there is a spatial mismatch between the two joints. Instead we shift the confidence map of the wrist towards the elbow through geometrical transform kernel and then add the transformed confidence map to that of the elbow. The high confidence map of the elbow is obtained finally.

 Although we can obtain all confidence maps of each body part, it is still uncertain whether these body parts belong to the same person or some joints with the same type among multiple people are overlapping. So some body parts may be incorrectly connected. Our network architecture is robust to shooting angles and occlusion similar to~\cite{Jain2013Learning}. Especially, in our network architecture we constrain the relationship among different body parts according to the BGSIM proposed in Section~\ref{bsim-sec} to achieve highly accurate location results, and BGSIM is proven to be highly robust to shooting angles and occlusion.
\subsection{BGSIM}
\label{bsim-sec}

\begin{figure}[t!]
\centering
\includegraphics[width=3.2in]{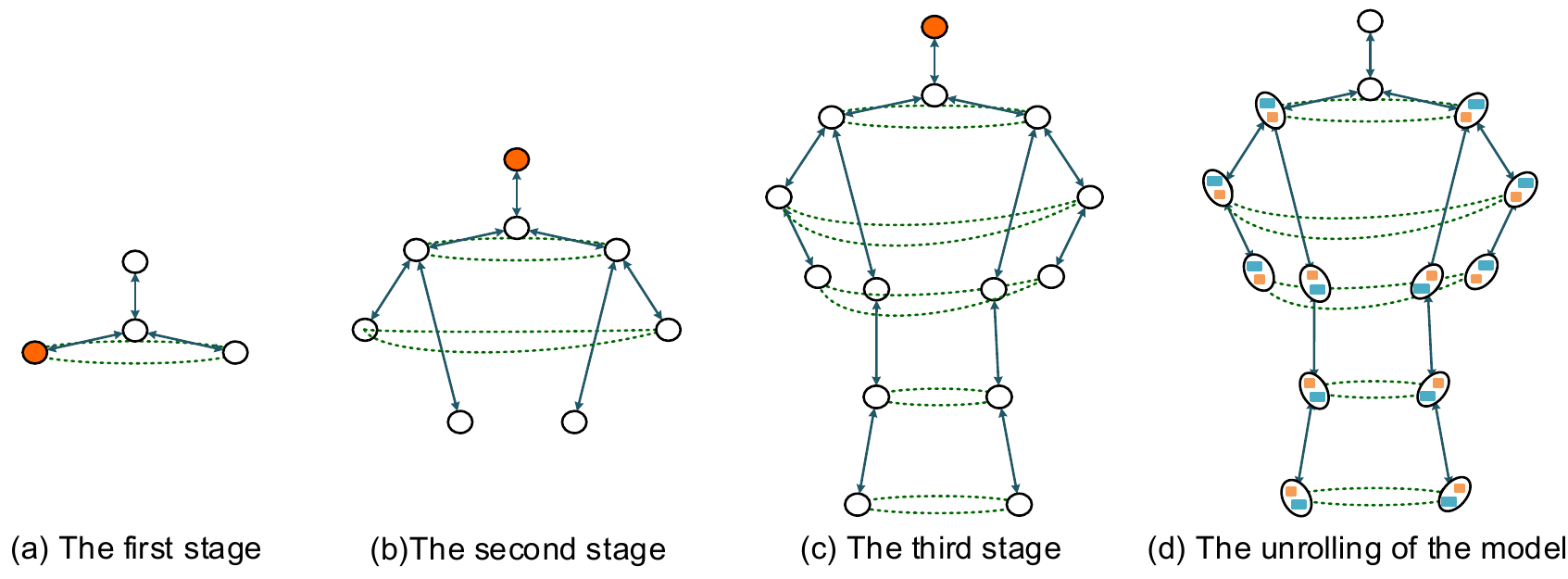}
\caption{The BGSIM at each stage and the unrolling of the model. (a) shows the BGSIM at the first stage, (b) and (c) show the BGSIM at the second and third stages, and (d) shows unrolling of the model}
\label{fig:5}
\end{figure}

\begin{figure}[t!]
\centering
\includegraphics[width=3.2in]{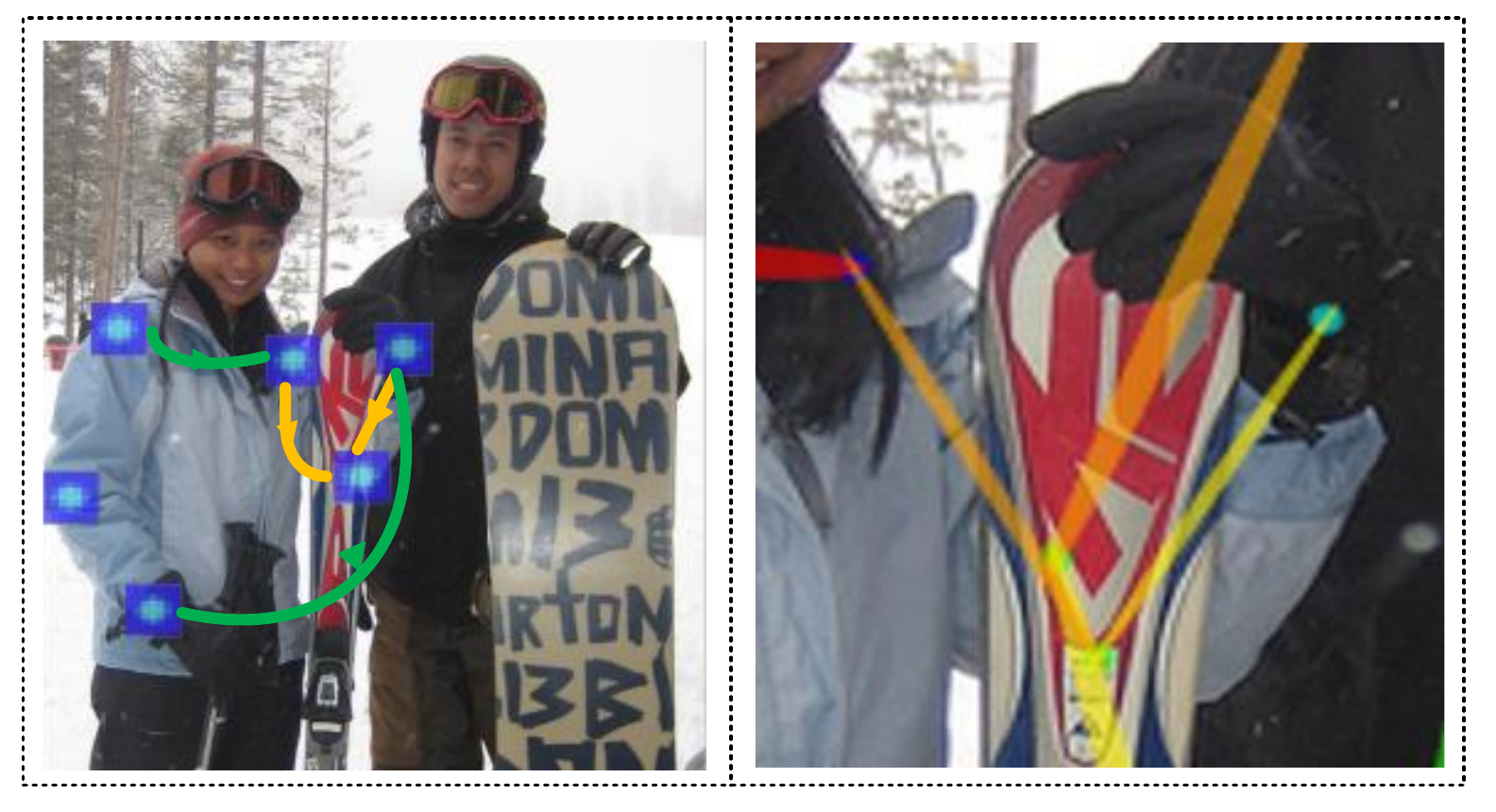}
\caption{Information propagation through BGSIM. In the left side of Figure~\ref{fig:BGSIM}, the left elbow of the woman is completely occluded by other thing; the right side of Figure~\ref{fig:BGSIM} is correct connected result}
\label{fig:BGSIM}
\end{figure}

 Considering the tree-structured model is simple and cannot solve occlusion problem, we propose BGSIM based on graph-structured model. At each stage, our proposed BGSIM selects the joint with the highest response of confidence maps as the base point (the orange joint point in Figure~\ref{fig:5}) in this stage. Then we use the base point as the root node of a series of tree structures unrolled by our BGSIM to propagate information to other joints, especially for the occluded joints. As is shown in Figure~\ref{fig:BGSIM}, the left elbow of the woman is completely occluded by other object, so we cannot capture its confidence map. However, The BGSIM is used to compute the confidence map of her left elbow by using the confidence maps of her left shoulder and left elbow. Meanwhile, the BGSIM propagate informations from her right shoulder and right wrist to her left shoulder and left wrist respectively, thereby getting richer information of her left shoulder and left wrist and computing more explicit position of the left elbow's confidence map. The right side of Figure~\ref{fig:BGSIM} is correct joint connection after utilizing BGSIM to compute the confidence map of the occluded joint.

In the BGSIM, we denote $G=(V,\varepsilon)$ be the $N$-node graph, where $V$ denotes the set of body parts, $\varepsilon$ denotes the set of constraint edges and $N=\left|V\right|$ is the number of joints. The position of the $i$ joint point is denoted as ${p_{i}}=\{{x_{i}},{y_{i}}\}$. The set of pairwise spatial relationships are denoted as $t=\{{t_{ij}},{t_{ji}}|(i,j)\in \varepsilon \}$, where ${t_{ij}},{t_{ji}}\in \{1,\cdot \cdot \cdot ,{T_{ij}}\}$ denotes the relative position between joint $i$ and $j$. $\phi=(p,t,o)$ is the pose configuration, $p = \{ {p_i}\}$ and $o = \{ {o_i}\}$, ${{o}_{i}}\in \{0,1,2\}$ is the occlusion state (``0'' for visibility, ``1'' for self-occlusion, and ``2'' for occlusion by other objects).We can encode richer information through bi-directional information propagation between joint i and joint.

Given an input image $I$, the goal of the BGSIM is to maximize the posterior as follows
\begin{equation}
\begin{split}
P(\phi |I)\propto (\sum\limits_{i\in V}{U(I,{{p}_{i}},{{o}_{i}})}+\sum\limits_{(i,j)\in {{\varepsilon }_{K}}}{{{R}^{K}}({{t}_{ij}},{{t}_{ji}},{{o}_{i}},{{o}_{j}})} \\
 +\sum\limits_{(m,n)\in {{\varepsilon }_{C}}}{{{R}^{C}}({{t}_{mn}},{{t}_{nm}},{{o}_{m}},{{o}_{n}})})
\end{split}
\end{equation}
where ${{\varepsilon }_{K}}$ denotes the kinematic constraints among body parts, ${{\varepsilon }_{C}}$ denotes the set of additional constraints among body parts that are not physically connected, and $\varepsilon ={{\varepsilon }_{K}}\cup {{\varepsilon }_{C}}$ denotes the full set of constraints. For simplicity, we call them kinetic edges (${{\varepsilon }_{K}}$) and contextual edges (${{\varepsilon }_{C}}$) respectively. In addition, $U(I,{{p}_{i}},{{o}_{i}})$ is the local part appearance score considering the occlusion state, while ${{R}^{K}}({{t}_{ij}},{{t}_{ji}},{{o}_{i}},{{o}_{j}})$ and ${{R}^{C}}({{t}_{mn}},{{t}_{nm}},{{o}_{m}},{{o}_{n}})$ represent the deformation (kinetic and contextual respectively) scores with occlusion coherence. The score function of BGSIM is formulated as:
\begin{equation}
S=\frac{F(I,\phi )}{\left| F(I,\phi ) \right|+\left| F'(I,\phi ) \right|}+\frac{F'(I,\phi )}{\left| F(I,\phi ) \right|+\left| F'(I,\phi ) \right|}
\end{equation}
$F(I,\phi )$ denotes score function with forward information flow (we define that the forward direction is from the base point to the end joint point of human body and the backward direction is opposite), $F'(I,\phi )$ denotes score function with backward information flow. The calculation of $F(I,\phi )$ and $F'(I,\phi )$ is same, so we only analyze the $F(I,\phi )$

\begin{equation}
\begin{split}
F(I,\phi )=\sum\limits_{i\in V}{U(I,{{p}_{i}},{{o}_{i}})}+\sum\limits_{(i,j)\in {{\varepsilon }_{K}}}{{{R}^{K}}({{t}_{ij}},{{t}_{ji}},{{o}_{i}},{{o}_{j}})} \\
+\sum\limits_{(m,n)\in {{\varepsilon }_{C}}}{{{R}^{C}}({{t}_{mn}},{{t}_{nm}},{{o}_{m}},{{o}_{n}})}
\end{split}
\end{equation}
the three terms on the right side of Equation (3) handle occlusion from different aspects. The first term considers occlusion ``local'', i.e., whether each local part is visible, other-occluded or self-occluded. The second term encourages the occlusion states among physically connected parts to be continuous, while the third term reflects the self-occlusion relationship between physically non-connected parts.

The part appearance score is represented as
\begin{equation}
U(I,{p_i},{o_i}) = {w_i}f(i|I({p_i});\theta ) \cdot {L_{\{ 0,1\} }}({o_i}) + {b_i}({o_i})
\end{equation}
where ${w_i}$ is appearance weight parameter, $f(i|I({p_i});\theta )$ is the image evidence based on the local image $I({p_i})$, $\theta$ denotes the parameters of convolutional neural network, ${L_{\{ 0,1\} }}({o_i})$ is an indicator function as below:
\begin{equation}
{L_{\{ 0,1\} }}({o_i}) = \left\{ \begin{array}{l}
1{\kern 1pt} ,{\kern 1pt} {\kern 1pt} {\kern 1pt} {\kern 1pt} {\kern 1pt} {\kern 1pt} {\kern 1pt} {\kern 1pt} {\kern 1pt} {\kern 1pt} {\kern 1pt} {\kern 1pt} {\kern 1pt} if{\kern 1pt} {\kern 1pt} {\kern 1pt} {o_i} = 0,1\\
0{\kern 1pt} ,{\kern 1pt} {\kern 1pt} {\kern 1pt} {\kern 1pt} {\kern 1pt} {\kern 1pt} {\kern 1pt} {\kern 1pt} {\kern 1pt} {\kern 1pt} {\kern 1pt} if{\kern 1pt} {\kern 1pt} {\kern 1pt} {o_i} = 2
\end{array} \right.
\end{equation}

The kinetic pairwise score is formulated as
\begin{equation}
\begin{split}
{R^K}(I,{p_i},{p_j},{t_{ij}},{t_{ji}},{o_i},{o_j}) = w_{ij}^{{t_{ij}}} \cdot \mu ({p_j} - {p_i} - r_{ij}^{{t_{ij}}})
\\
+ {w_{ij}}f({t_{ij}}|I({p_i});\theta ) \cdot {L_{\{ 0,1\} }}({o_i})
\\
+ w_{ji}^{{t_{ji}}} \cdot \mu ({p_i} - {p_j} - r_{ji}^{{t_{ji}}})
\\
+ {w_{ji}}f({t_{ji}}|I({p_j});\theta ) \cdot {L_{\{ 0,1\} }}({o_j})
\\
+ O_{ij}^{{t_{ij}}{t_{ji}}}({o_i},{o_j})
\end{split}
\end{equation}
where $u (\Delta p = [\Delta x - \Delta y]) = {[\Delta x,\Delta {x^2},\Delta y,\Delta {y^2}]^T}$ are the deformation features, $r_{ij}^{{t_{ij}}}$ is the mean relative position of special relationship type ${t_{ij}}$. $w_{ij}^{{t_{ij}}}$, ${w_{ij}}$, $w_{ji}^{{t_{ji}}}$, ${w_{ji}}$ are the weight parameters of special relationship types, and $O_{ij}^{{t_{ij}}{t_{ji}}}({o_i},{o_j})$ is the bias term which encodes the occlusion coherence between part $i$ and part $j$.

The contextual pairwise score is formulated as
\begin{equation}
\begin{split}
{R^C}(I,{p_m},{p_n},{t_{mn}},{t_{nm}},{o_m},{o_n}) = w_{mn}^{{t_{mn}}} \cdot \mu ({p_n} - {p_m} - r_{mn}^{{t_{mn}}})
\\
+ w_{nm}^{{t_{nm}}} \cdot \mu ({p_m} - {p_n} - r_{nm}^{{t_{nm}}})
\\
+ O_{mn}^{{t_{mn}}{t_{nm}}}({o_m},{o_n})
\end{split}
\end{equation}
where all the terms on the right hand side are in accordance with that in Equation (6). The bias term $O_{mn}^{{t_{mn}}{t_{nm}}}({o_m},{o_n})$ reflects long range spatial constraints as well as occlusion between non-adjacent parts m and n.
\begin{figure}[t!]
\centering
\includegraphics[width=3.2in]{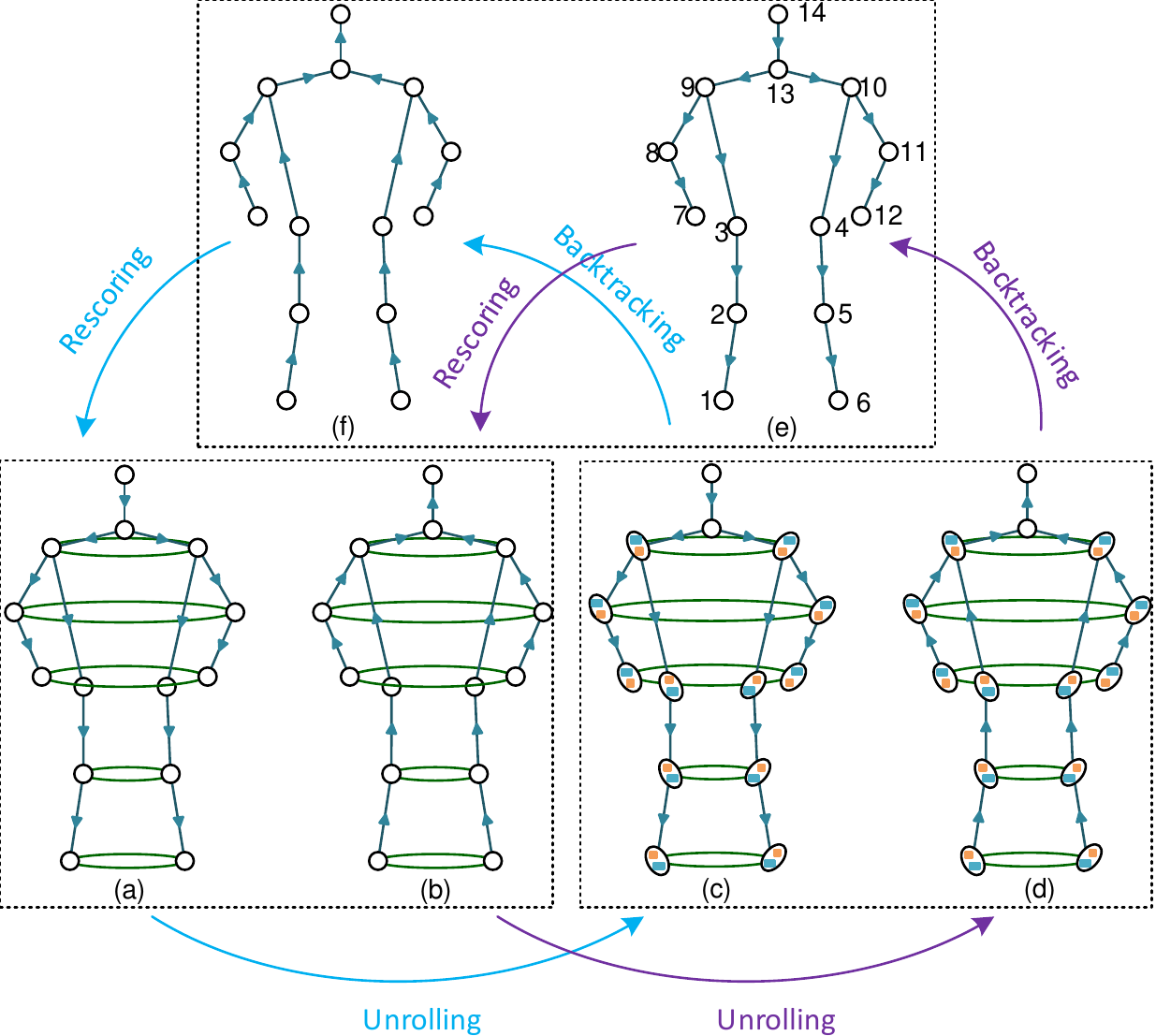}
\caption{Inference of BGSIM. (a) BGSIM in downward direction, (b) BGSIM in upward direction, (c) and (d) unrolled computation tree, and (e) and (f) nodes backtracked via the tree structure}
\label{fig:6}
\end{figure}

\begin{figure*}[t!]
\centering
\includegraphics[width=6.7in]{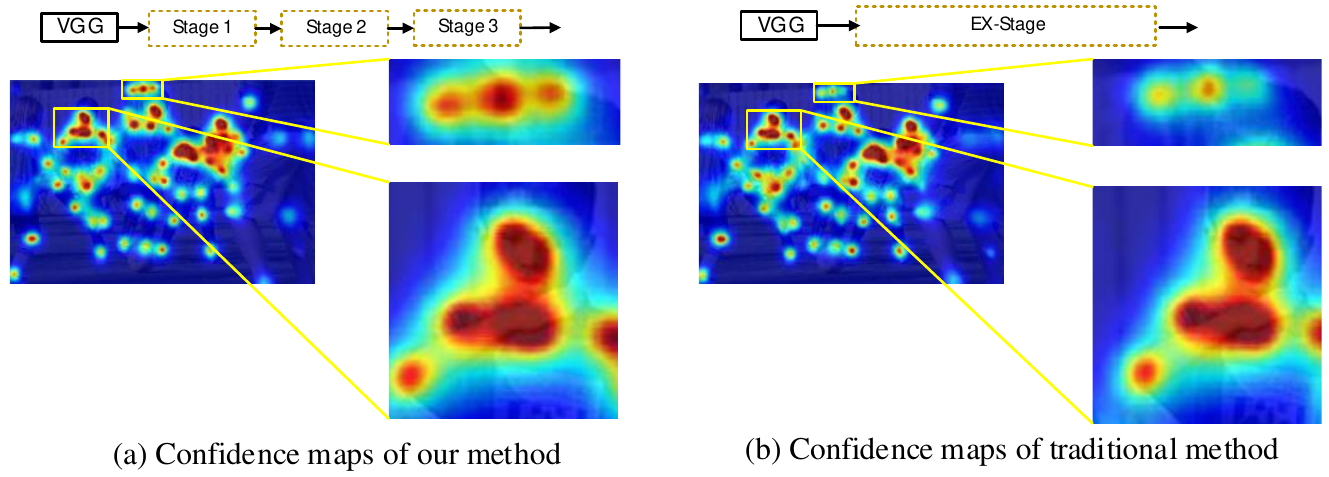}
\caption{The response of some occluded joints is strengthened by optimizing the three stages, and our method makes the response highly centralized and accurate}
\label{fig:8}
\end{figure*}

In fact, we unroll the BGSIM formed at each stage into a tree model to generate the pose hypothesis and then rescore the candidate pose configurations with BGSIM. Afterward, we introduce the inference of the proposed model in the forward information flow and the inference of the proposed model in the backward information flow follows the same operation.

We simply the symbols of the generic score function in Equation (3) as follows
\begin{equation}
F(I,\phi ) = \sum\limits_{i \in V} {{U_i}(\phi ) + \sum\limits_{(i,j) \in {\varepsilon _K}} {R_{ij}^K(\phi ) + \sum\limits_{(i,j) \in {\varepsilon _C}} {R_{ij}^C(\phi )} } }
\end{equation}

\textbf{1) Model Unrolling:} for any part $i$ with an out-degree(number of connections pointing to the other parts) of ${{\nu }_{i}}>1$, we generate ${{\nu }_{i}}-1$  virtual parts and unroll the contextual edges to form a computation tree similar to~\cite{Tatikonda2002Loopy} as shown in Figure~\ref{fig:6} (d). The unrolled tree model is used to generate pose configurations.

\textbf{2) Pose Selection:} The goal of our graphical model is to maximize the score in Equation (8), i.e.,
\begin{equation}
{\phi _m} = \mathop {\arg \max }\limits_{X'} \sum\limits_{i \in V} {{U_i}(\phi ) + \sum\limits_{(i,j) \in {\varepsilon _K}} {R_{ij}^K(\phi ) + \sum\limits_{(i,j) \in {\varepsilon _C}} {R_{ij}^C(\phi )} } }
\end{equation}
where ${\phi _m}$ is the optimal pose configuration with the proposed BGSIM. Instead of propagating information on a loopy graph, we propagate information on the unrolled tree structure to generate root hypothesis. This allows us to employ dynamic programming to propagate information from leaf nodes to the root node efficiently.

\begin{figure*}[t!]
\centering{}
\includegraphics[width=6.7in]{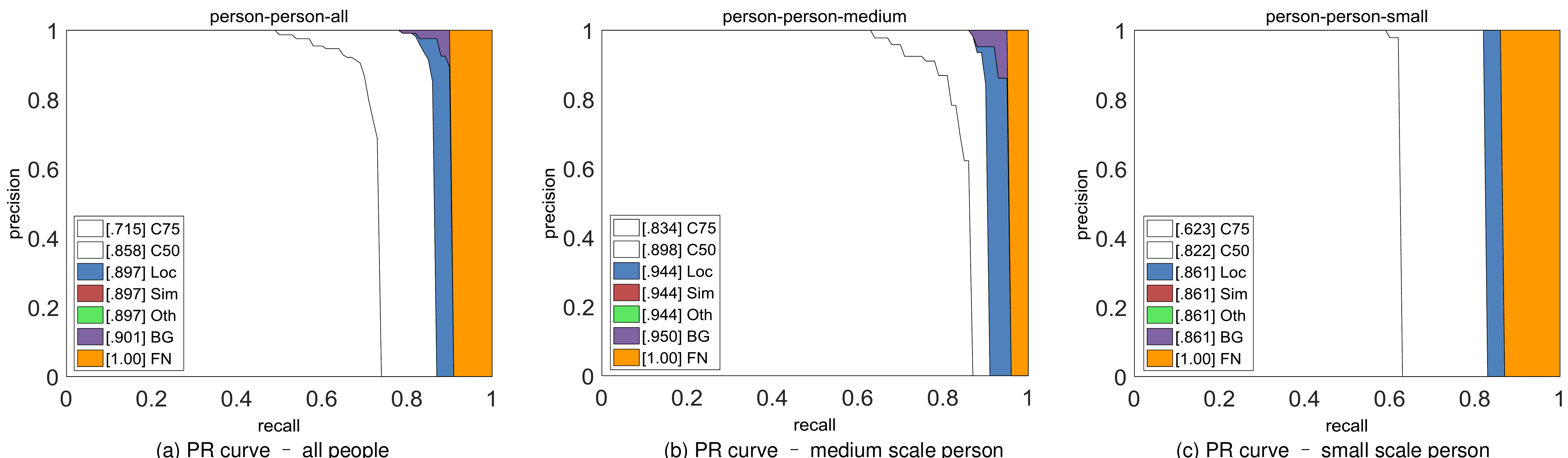}
\caption{ AP performance on the COCO validation set in (a), (b), and (c)}
\label{fig:9}
\end{figure*}

The optimization over the unrolled model is formulated as
\begin{equation}
\begin{aligned}
{\phi'_m} = &\mathop {\arg \max }\limits_{X'} \left\{\sum\limits_{i \in V} {{U_i}(\phi )} + \sum\limits_{m \in V'} {{U_k}(\phi )}+\right.\\
&\left. \sum\limits_{(i,j) \in {\varepsilon _K}} {R_{ij}^K(\phi )}+\sum\limits_{(m,n) \in {\varepsilon _C}} {R_{ij}^C(\phi )}\right\}
\end{aligned}
\end{equation}
where ${\phi'_m}$ is the optimal pose configuration of the unrolled tree-structured model. This is equivalent to adding part appearance weights to those nodes that have more connections.

Suppose the number of possible root hypotheses to be $H$ in the test image. We sort them according to root score and select top ${H_{\sigma}}/{H}$ is the ratio of hypothesis selection. And we assume that the optimal hypothesis is included in the selected top-$\sigma$ hypotheses of the unrolled configurations.

\textbf{3) Backtracking and Rescoring:} As soon as the top-$\sigma$ hypotheses of root nodes are determined, the corresponding pose configurations can be obtained by backtracking directly from the root node to the leaf nodes. We only backtrack the child node from the real parent node (e.g., node 1 is backtracked from node 2 rather than node 6 in Figure~\ref{fig:6} (e)) because the parent near the root node has less freedom of movement and is more reliable. We recompute the score of the pose configurations with a graphical model and rerank the hypothesis to obtain the optimal pose configuration.

\begin{table}[t!]
\renewcommand{\arraystretch}{1.3}
\begin{center}
\caption{Results on the COCO dataset. $A{P^{50}}$is the AP at OKS=0.5, and $A{P^{L}}$ is the AP for a large-scale person}
\label{table:coco}
\begin{tabular}{|l|l|l|l|l|l|}
\hline
{} & $A{P^{}}$ & $A{P^{50}}$ & $A{P^{75}}$ & $A{P^{M}}$ & $A{P^{L}}$ \\ \hline
\hline
CMU-Pose & 61.8 & 84.9 & 67.5 & 57.1 & 68.2 \\
\hline
G-RMI & 60.5 & 82.2 & 66.2 & 57.6 & 66.6 \\
\hline
DL-61 & 54.4 & 75.3 & 50.9 & 58.3 & 54.3 \\
\hline
R4D & 51.4 & 75.0 & 55.9 & 47.4 & 56.7 \\
\hline
umich\_v1 & 46.0 & 74.6 & 48.4 & 38.8 & 55.6 \\
\hline
Caltech & 40.2 & 65.2 & 41.9 & 34.9 & 49.2 \\
\hline
RMPE & 61.8 & 83.7 & $\bf{69.8}$ & $\bf{58.6}$ & 67.6 \\
\hline
\textbf{Ours} & $\bf{62.9}$ & $\bf{85.1}$ & 69.7 & 56.7 & $\bf{70.2}$ \\
\hline
\end{tabular}
\end{center}
\end{table}

\section{Experimental Results}

Our model is only trained on the COCO 2014 dataset and evaluated on two public benchmarks for person pose estimation, namely, the (1) COCO 2014 dataset and the (2) MPII human multi-person dataset. These two datasets contain many real-world challenging scenes, such as crowding, occlusion, and contact. Our approach demonstrates the same performance and even outperforms the other state-of-the-art methods on the COCO and MPII datasets, and shows its best performance in our own multi-person ``occlusion" dataset with different levels of occlusion.

In our method, the first stage predicts the heads and shoulders, the second stage predicts the elbows and hips, and the third stage predicts the wrists, knees, and ankles. Each stage shares the output information of the previous stage as the input to optimize the joints at this stage. Our method shows a very significant improvement in responding to some partially or severely occluded joints in a cascading way. Compared with traditional CNN architecture (Figure~\ref{fig:8} (b)), the response of occluded joints in our architecture is strengthened by optimizing the three stages, and our method makes the response highly centralized and accurate. In our experiment, we use the convolutions with a kernel size of $7*7$. In fact, the larger kernel size has a larger receptive field, that can obtain rich global information, which is especially helpful when dealing with occluded joints.

\subsection{Results on the COCO dataset}
The training, validation, and testing sets of the COCO dataset contain over 200K images and 250K persons-instances-points labels with keypoints. Each person is annotated with 15 body joints for the full body. We apply the following object keypoint similarity (OKS)~\cite{Lin2014Microsoft} defined by the COCO evaluation:
\begin{equation}
\ OKS = \frac{{\sum\limits_i {\left[ {\exp \left( { - \frac{{d_i^2}}{{2{s^2}k_i^2}}} \right)\delta \left( {{v_i} > {\rm{0}}} \right)} \right]} }}{{\sum\limits_i {\left[ {\delta \left( {{v_i} > {\rm{0}}} \right)} \right]} }}\
\end{equation}

\begin{figure}[t!]
\centering
\includegraphics[width=3.3in]{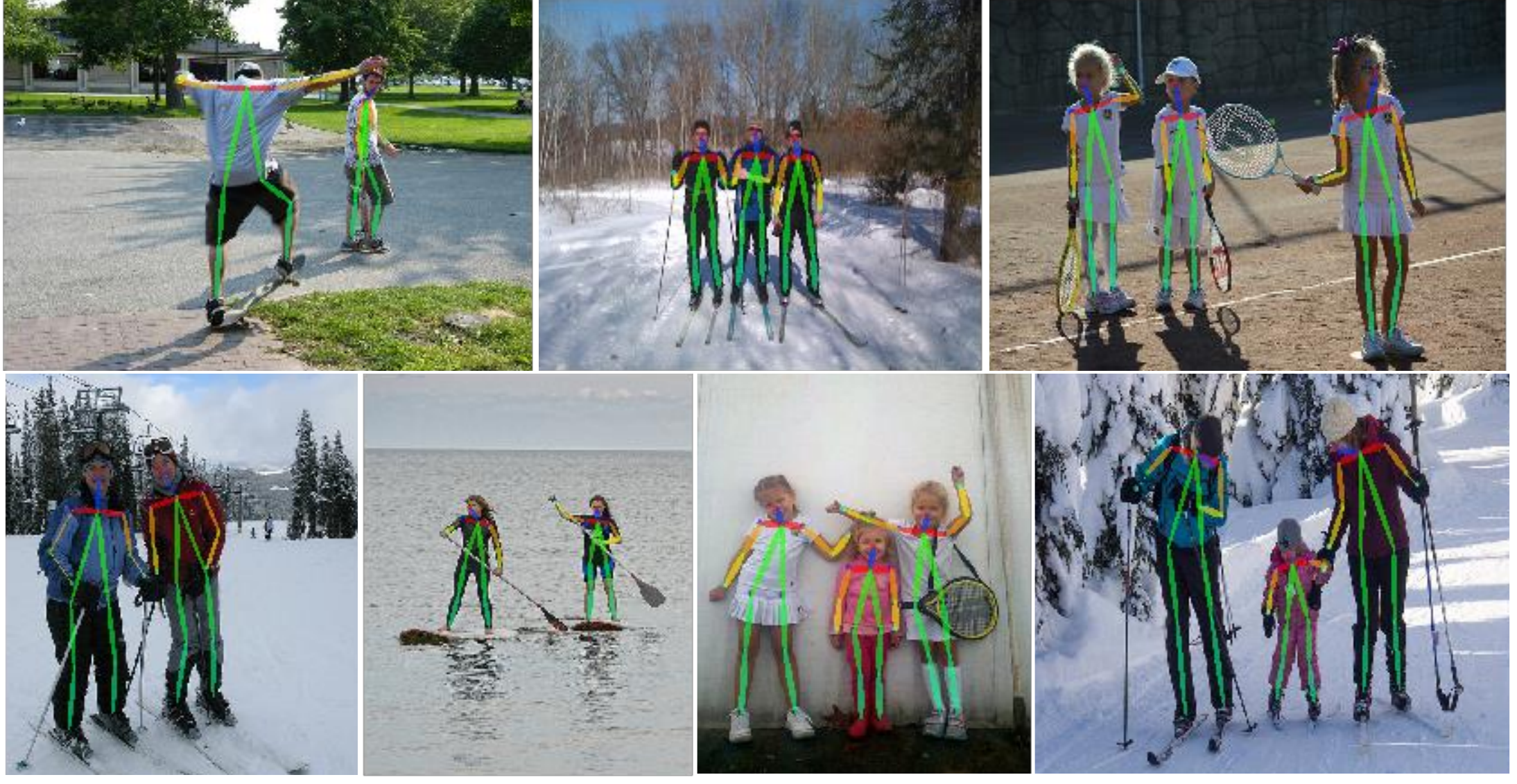}
\centering
\caption{Experimental visualization results on COCO dataset}
\label{fig:COCO}
\end{figure}

The OKS is analogous to the IoU in object detection. By using OKS, we can compute the mean average precision (AP) and average recall (AR) as competing metrics. Table~\ref{table:coco} shows the mAP of our method with different OKS thresholds and compares the mAP performance of our method with that of other approaches. For $A{P^{50}}$, our method outperforms the other methods and achieves an 85.1\char`\%{} AP. For $A{P^{M}}$, our architecture shows a slightly poorer performance than state-of-the-art architecture. Our architecture shows the best performance at $A{P^{L}}$ and achieves 70.2\char`\%{} AP. Figure~\ref{fig:9} shows a breakdown of the errors of our method on the COCO validation set. The area under each curve is shown in brackets in the legend. C75 denotes PR(precision recall) at IoU(Intersection over Union)=.75, and area under curve corresponds to ${AP}^{IoU=.75}$. Loc represents PR at IoU=.10(localization errors ignored, but not duplicate detections) and all remaining settings use IoU=.1. Sim denotes PR after supercategory false positives (fps) are removed and Oth represents PR after all class confusions are removed. BG denotes PR after all background (and class confusion) fps are removed and FN represents PR after all remaining errors are removed (trivially AP=1). In the case of all people, AP at IoU=.75 is .715 and perfect localization would increase AP to .897. Interesting, removing all class confusions (both within supercategory and across supercategories) would not raise AP. Removing background fp would only raise AP slightly to .901. In summary, Most of the false positive results are dominated by imperfect 
localization rather than background confusion, thereby indicating a large space for improvement in capturing spatial dependencies than in recognizing the appearance of body parts.  Figure~\ref{fig:COCO} shows some experimental results on COCO dataset.

\begin{figure}[t!]
\centering
\includegraphics[width=3.3in]{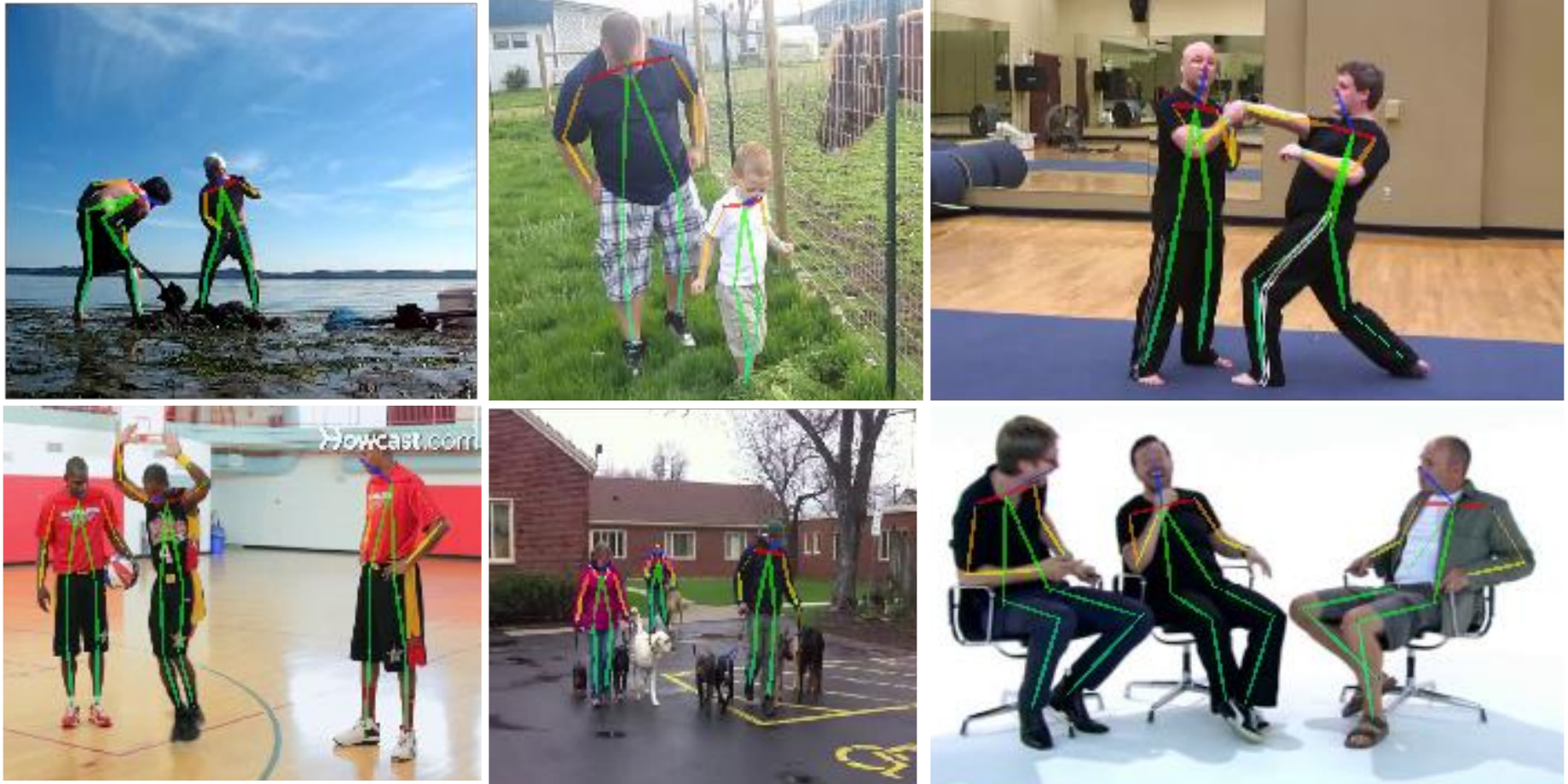}
\centering
\caption{Experimental visualization results on MPII dataset}
\label{fig:MPII}
\end{figure}


\begin{table*}[t!]
\renewcommand{\arraystretch}{1.8}
\begin{center}
\caption{Results on the MPII dataset}
\label{table:MPII}
\begin{tabular}{|c|c|c|c|c|c|c|c|c|}
\hline
{} & Head & Shoulder & Elbow & Wrist & Hip & Knee & Ankle & Total \\
\hline
\hline
Iqbal\&Gall, ECCV16~\cite{Iqbal2016Multi} & 58.4 & 53.9 & 44.5 & 35.0 & 42.2 & 36.7 & 31.1 & 43.1 \\
\hline
DeeperCut, EVVC16~\cite{Insafutdinov2016DeeperCut} & 78.4 & 72.5 & 60.2 & 51.0 & 57.2 & 52.0 & 45.4 & 59.5 \\
\hline
Levinkov et al., CVPR17~\cite{Levinkov2016Joint} & 89.8 & 85.2 & 71.8 & 59.6 & 71.1 & 63.0 & 53.5 & 70.6 \\
\hline
Insafutdinov et al., CVPR17~\cite{Eldar2016ArtTrack}
 & 88.8 & 87.0 & 75.9 & 64.9 & 74.2 & 68.8 & 60.5 & 74.3 \\
 \hline
CMU-Pose~\cite{Cao2016Realtime} & $\bf{91.2}$ & 87.6 & 77.7 & 66.8 & $\bf{75.4}$ & 68.9 & 61.7 & 75.6 \\
\hline
RMPE~\cite{Fang2016RMPE} & 88.4 & 86.5 & 78.6 & 70.4 & 74.4 & $\bf{73.0}$ & 68.5 & 76.7 \\
\hline
\textbf{Ours} & 90.7 & $\bf{87.7}$ & $\bf{80.2}$ & $\bf{71.1}$ & 74.8 & 72.9 & $\bf{66.3}$ & $\bf{77.6}$ \\
\hline
\end{tabular}
\end{center}
\end{table*}

\begin{figure*}[t!]
\centering
\includegraphics[width=6.7in]{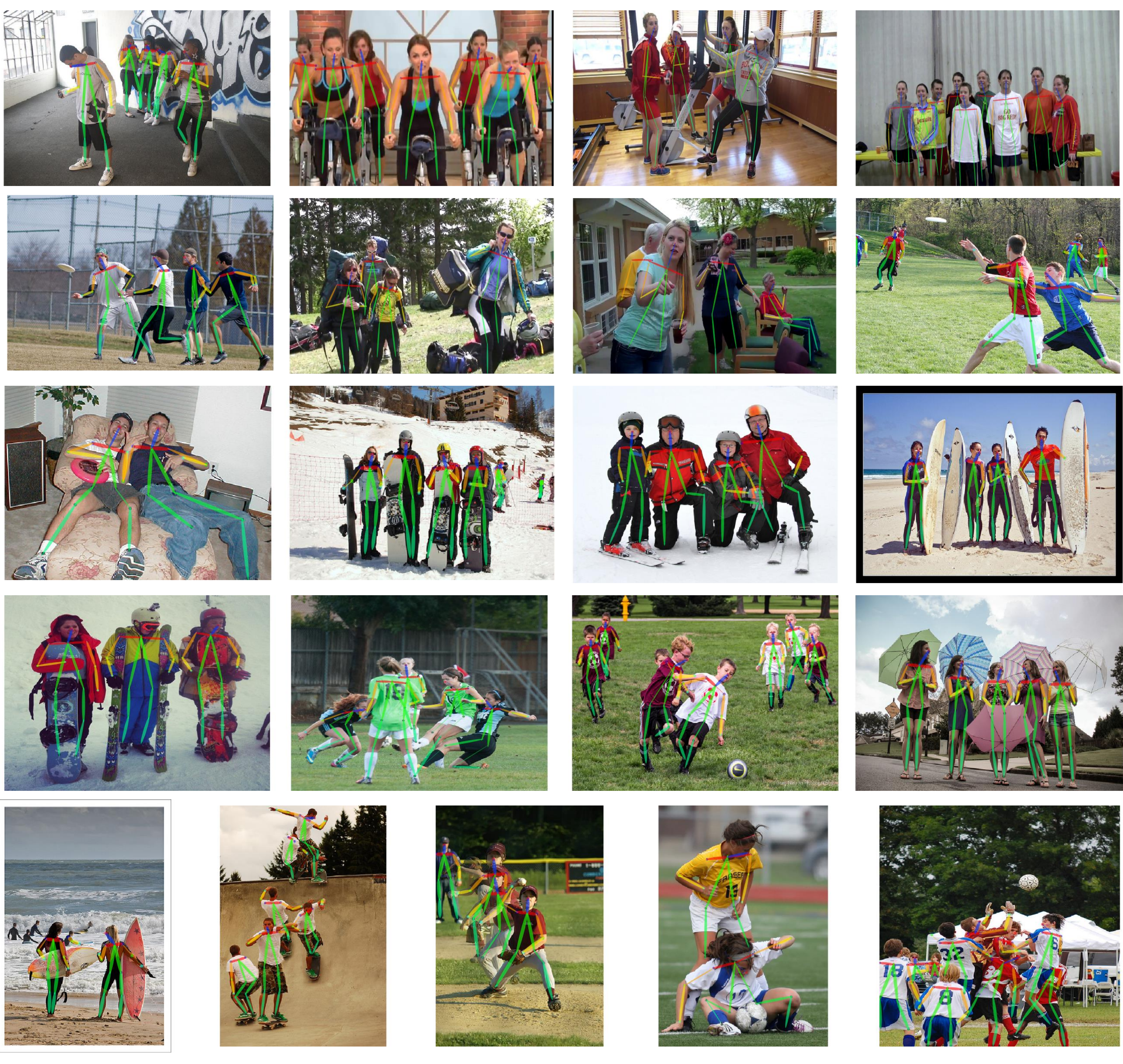}
\centering
\caption{Experimental results on our own dataset, which images are selected from the AI challenger, COCO, MPII, and LSP public datasets}
\label{fig:10}
\end{figure*}

\begin{table*}[t!]
\renewcommand{\arraystretch}{1.3}
\begin{center}
\caption{Results on our own dataset}
\label{table:3}
\begin{tabular}{|p{1.5cm}<{\centering}|p{1cm}<{\centering}|p{1cm}<{\centering}|p{1cm}<{\centering}|p{1cm}<{\centering}|p{1cm}<{\centering}|p{1cm}<{\centering}|p{1cm}<{\centering}|p{1cm}<{\centering}|}
\hline
{} & $A{P^{}}$ & $A{P^{50}}$ & $A{P^{75}}$ & $A{P^{M}}$ & $A{P^{L}}$ & $A{R^{}}$ & $A{R^{50}}$ & $A{R^{75}}$\\ \hline
\hline
CMU-Pose & 39.7 & 57.3 & 41.9 & 36.2 & 42.6 & 40.6 & 49.1 &41.2 \\
\hline
Ours & $\bf{41.8}$ & $\bf{58.6}$ & $\bf{42.6}$ & $\bf{37.9}$ & 42.1 & $\bf{41.3}$ & 48.8 & $\bf{42.2}$ \\
\hline
\end{tabular}
\end{center}
\end{table*}

\subsection{Results on the MPII dataset }

The MPII Human Pose dataset includes around 25000 images containing over 40000 persons with annotated body joints. We use the toolkit provided in~\cite{Pishchulin2015DeepCut} to measure the Average Precision (mAP) of all body parts based on the PCKh threshold. Table~\ref{table:MPII} compares the mAP performance of our method with that of other state-of-the-art methods. 
Compared with~\cite{Cao2016Realtime} and RMPE, our method shows the similar performance for some body parts, such as heads or shoulders. However, for other body parts that are easily occluded or show a large variation in the limb orientation (e.g., elbows or wrists), our method outperforms these two state-of-the-art methods. Unlike top-down methods such as RMPE, our method does not rely on human detection. Our method also shows the same performance as RMPE for single-person detection and can effectively detect multiple targets. Figure~\ref{fig:MPII} shows some experimental results on MPII dataset.

\subsection{Results on our own dataset}

Our dataset contains 5899 multi-person and high occlusion images that are selected from the AI challenger~\cite{Wu2017AI}, COCO, MPII, and LSP public datasets. We directly evaluate the new dataset ``occlusion'' without any extra training. Figure~\ref{fig:10} shows some experimental results on our own dataset.
Table~\ref{table:3} compares our results on our own dataset with those of CMU-Pose~\cite{Cao2016Realtime}. Average Precision (AP) is primary challenge metric, the performance of the Ours reaches 41.8\char`\%{} $A{P^{}}$, which is 2.1\char`\%{} higher than that reported in~\cite{Cao2016Realtime}. $A{P^{75}}$ represents strict metric, our method achieves 42.6\char`\%{}. $A{P^{M}}$ is $A{P^{}}$ for medium scale person and $A{P^{L}}$ is $A{P^{}}$ for large scale person. From Table~\ref{table:3}, we can see our method outperforms CMU-Pose for medium scale person. For $A{R^{}}$ (Average Recall) and $A{R^{75}}$, the performance of our method is also higher than CMU-Pose. Thus, our method has a significant application value especially for pose estimation with high occlusions.

\section{Conclusion}
In this paper, we follow the bottom-up pipeline for multi-person pose estimation and a new bi-directional graph structure information model(BGSIM) is presented to address the occluded keypoints. More specifically, our proposed BGSIM is involved into a multi-stage network architecture to encode rich contextual information among all joints. In addition, we dynamically the base point with highest reponse of confidence maps in BGSIM and take it as root node to propagate information to other joints, especially for the occluded joints, via tree stuctures which is unrolled by our proposed BGSIM. Our method achieves state-of-the-art results on the COCO keypoint and MPII benchmark datasets as well as achieves the best results on our own selected dataset without any extra training.
%


%



%
%

\ifCLASSOPTIONcaptionsoff
  \newpage
\fi




\bibliographystyle{./IEEEtran}
\bibliography{./egbib}
%


%
\begin{IEEEbiography}[{\includegraphics[width=1in,height=1.25in,clip,keepaspectratio]{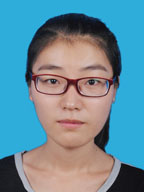}}]{Jing Wang}
is a graduate student in the School of Information Engineering of Zhengzhou University, China. Her research interest is computer graphics and computer vision, particulary in deep learning.
\end{IEEEbiography}
\vspace{-12 mm}

\begin{IEEEbiography}[{\includegraphics[width=1in,height=1.25in,clip,keepaspectratio]{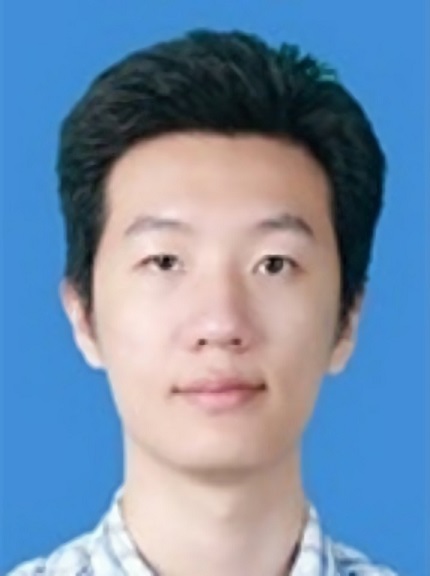}}]{Ze Peng}
 is a M.S. in the Industrial Technology Research Institute of Zhengzhou University, China, and his research interest is
 deep learning and computer vision.
\end{IEEEbiography}
\vspace{-10 mm}

\begin{IEEEbiography}[{\includegraphics[width=1in,height=1.25in,clip,keepaspectratio]{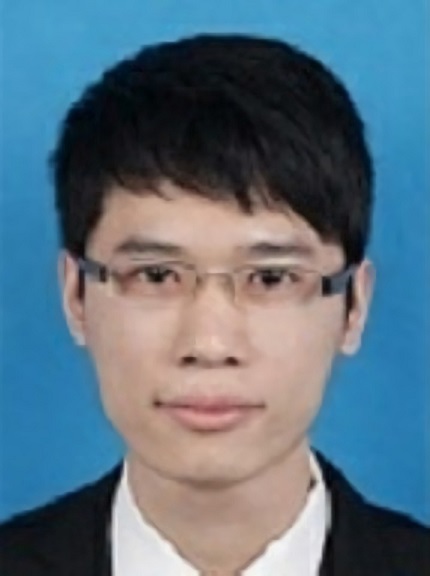}}]{Junyi Sun}
is a M.S. in the Industrial Technology Research Institute of Zhengzhou University, China, and his research interest is
 deep learning and computer vision.
\end{IEEEbiography}

\vspace{-10 mm}

\begin{IEEEbiography}[{\includegraphics[width=1in,height=1.25in,clip,keepaspectratio]{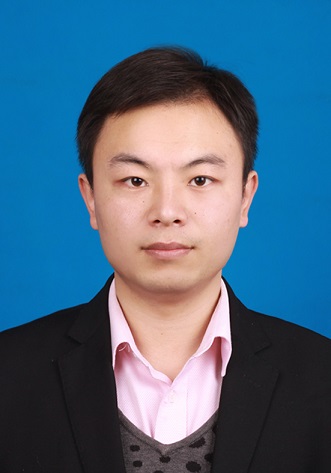}}]{Pei Lv}
is an associate professor in School of Information Engineering, Zhengzhou University, China. His research interests include video analysis and crowd simulation. He received his Ph.D in 2013 from the State Key Lab of CAD\&CG, Zhejiang University, China. He has authored more than 20 journal and conference papers in these areas, including IEEE TIP, IEEE TCSVT, ACM MM, etc.
\end{IEEEbiography}
\vspace{-10 mm}

\begin{IEEEbiography}[{\includegraphics[width=1in,height=1.25in,clip,keepaspectratio]{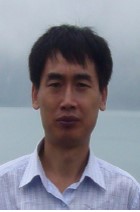}}]{Bing Zhou}
 is currently a professor at the School of Information Engineering, Zhengzhou University, Henan, China. He received the B.S. and M.S. degrees from Xian Jiao Tong University in 1986 and 1989, respectively,and the Ph.D. degree in Beihang University in 2003, all in computer science. His research interests cover video processing and understanding, surveillance, computer vision, multimedia applications.
\end{IEEEbiography}
\vspace{-10 mm}

\begin{IEEEbiography}[{\includegraphics[width=1in,height=1.25in,clip,keepaspectratio]{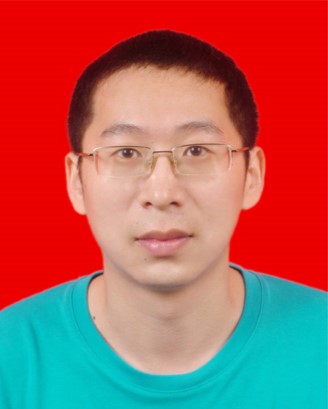}}]{Mingliang Xu}
 is a full professor in the School of Information Engineering of Zhengzhou University, China, and currently is the director of CIISR (Center for Interdisciplinary Information Science Research) and the vice General Secretary of ACM SIGAI China. He received his Ph.D. degree in computer science and technology from the State Key Lab of CAD\&CG at Zhejiang University, Hangzhou, China. He previously worked at the department of information science of NSFC (National Natural Science Foundation of China), Mar.2015-Feb.2016. His current research interests include computer graphics, multimedia and artificial intelligence. He has authored more than 60 journal and conference papers in these areas, including ACM TOG, ACM TIST, IEEE TPAMI, IEEE TIP, IEEE TIP, IEEE TCYB, IEEE TCSVT, ACM SIGGRAPH (Asia), ACM MM, ICCV, etc.
\end{IEEEbiography}

%
%
%

%
%
%




\end{document}